\documentclass[10pt]{article}

\usepackage[letterpaper,margin=0.82in]{geometry}
\usepackage[T1]{fontenc}
\usepackage[utf8]{inputenc}
\usepackage{lmodern}
\usepackage{microtype}
\usepackage{amsmath,amssymb}
\usepackage{booktabs}
\usepackage{enumitem}
\usepackage{xcolor}
\usepackage{graphicx}
\usepackage{tikz}
\usetikzlibrary{arrows.meta,positioning}
\usepackage[
  colorlinks=true,
  linkcolor=black,
  citecolor=blue!55!black,
  urlcolor=blue!55!black,
  pdfencoding=auto,
  psdextra
]{hyperref}

\hypersetup{
  pdftitle={Activation-Guided Graph Retrieval for Cognitive Memory Reconstruction in Language Model Systems},
  pdfauthor={Marvin Danig},
  pdfsubject={Architecture and analytic specification for activation-guided memory retrieval},
  pdfkeywords={AI memory, graph retrieval, activation, ACT-R, retrieval-augmented generation, evidence graphs}
}

\setlist{nosep,leftmargin=*}
\title{\textbf{Activation-Guided Graph Retrieval for\\Cognitive Memory Reconstruction in Language Model Systems}}
\author{Marvin Danig\\
\small Achiral Research\\
\small \href{mailto:marvin@achiral.ai}{\texttt{marvin@achiral.ai}}}
\date{\small Architecture and analysis preprint\\November 2026}

\begin{document}
\maketitle

\begin{abstract}
Retrieval-augmented language model systems generally select context by query-conditioned relevance. We argue that long-lived AI systems require a second operator: activation-conditioned memory availability. We specify an architecture in which request and goal cues seed an authorized memory graph, activation and fan attenuation allocate a bounded traversal frontier, retrieved entities recursively emit cues, and a typed evidence graph is reconstructed before generation. The specification yields five architectural consequences about candidate selection, traversal control, graph regularization, evidence reconstruction, and reinforcement. We derive implementation-level cost estimates and define ablations, failure conditions, and a reproducible evaluation protocol. The central claim is narrowed or falsified if a simpler graph reranker matches full traversal on causal and temporal recovery under equal budgets. This architecture-and-analysis preprint reports no benchmark measurements. If the claim holds, retrieval returns the evidence path that makes a memory available, not only a ranked list.
\end{abstract}

\section{Introduction}

Retrieval-augmented generation (RAG) gives language models access to information outside their parameters \cite{lewis2020rag}. Dense, sparse, hybrid, hierarchical, and graph-based retrievers have substantially improved candidate selection and corpus-level synthesis \cite{edge2024graphrag,sarthi2024raptor}. Persistent agent architectures additionally maintain records across interactions \cite{park2023generative,packer2023memgpt,rasmussen2025zep}. Yet persistence and retrieval do not by themselves define memory availability.

Let $M=\{m_1,\ldots,m_n\}$ be stored memory objects and $q$ a request. A conventional retriever ranks objects using a query-conditioned score
\begin{equation}
s_q(m_i)=\operatorname{sim}(f(q),f(m_i)),
\end{equation}
where $f$ is an embedding function and $\operatorname{sim}$ a similarity measure, possibly fused with lexical, graph, metadata, or learned-reranking features. This score answers which objects are relevant to $q$. It does not fully specify whether an object should influence inference now: whether it is current or superseded, repeatedly useful or merely popular, direct evidence or model-generated reflection, aligned with the active goal, contradicted by later evidence, or inaccessible under policy.

The motivating case is emergent memory, in which useful context forms through repeated work, reinforcement, forgetting, and revision rather than being stored by hand \cite{danig2026emergent}; the architecture applies to long-lived memory graphs in general.

We introduce \emph{activation-guided graph retrieval} (AGGR) as a stateful control layer over candidate generation and graph traversal. We call its output \emph{cognitive memory reconstruction}: rebuilding, before generation, a typed evidence graph of the experience that should shape the answer (Section~\ref{sec:reconstruction}). Its target is
\begin{equation}
P(m_i\text{ influences inference now}\mid q,g,H_t,G,\pi,\sigma_i),
\end{equation}
where $g$ is goal state, $H_t$ is interaction and use history, $G$ is graph structure, $\pi$ is authorization policy, and $\sigma_i$ is the epistemic state of memory $m_i$. Section~\ref{sec:activation} operationalizes this target with a runtime score $A_i$, an auditable ranking proxy for this probability rather than a calibrated estimate of it.

\paragraph{Contributions.}
This paper (i) formalizes the activation-versus-querying distinction (Section~\ref{sec:distinction}); (ii) specifies activation-guided, fan-aware graph traversal with recursive cue generation (Sections~\ref{sec:activation} and~\ref{sec:fan}); (iii) defines evidence-graph reconstruction as the retrieval output (Section~\ref{sec:reconstruction}); (iv) derives bounded-frontier, context, storage, and reinforcement estimates (Section~\ref{sec:estimates}); and (v) gives testable hypotheses, ablations, and failure conditions (Sections~\ref{sec:evaluation} and~\ref{sec:limits}).

\section{Activation Versus Querying}\label{sec:distinction}

Querying and activation are complementary operators over different conditionals. Query relevance can nominate candidates. Activation determines availability under system state.

\begin{table}[h]
\centering
\small
\begin{tabular}{p{0.18\linewidth}p{0.35\linewidth}p{0.35\linewidth}}
\toprule
\textbf{Dimension} & \textbf{Query-conditioned retrieval} & \textbf{Activation-conditioned memory} \\
\midrule
Question & What matches or answers $q$? & What experience should influence inference now? \\
State & Commonly request-local & Conditioned on goal, history, graph, policy, and epistemic state \\
Selection unit & Passage, document, row, node & Entity, episode, procedure, edge, path, or subgraph \\
Time & Usually static between index updates & Strengthens, decays, and changes with outcomes \\
Structure & Items or neighborhoods & Budgeted traversal over evidence paths \\
Output & Ranked context items & Typed, provenance-bearing evidence graph \\
\bottomrule
\end{tabular}
\caption{Querying estimates request relevance; activation controls memory availability.}
\label{tab:distinction}
\end{table}

The distinction is especially relevant for causal, temporal, procedural, and decision-lineage questions. For ``Why was Project X canceled?'', lexical or dense retrieval may return the cancellation announcement and semantically similar status reports. A memory system may instead need to reconstruct
\begin{equation}
\text{account loss}\rightarrow\text{forecast revision}\rightarrow
\text{budget reduction}\rightarrow\text{cancellation}.
\end{equation}
The decisive evidence can be several graph steps away and need not resemble the query. The relevant retrieval unit is a path under a goal-conditioned policy, not necessarily a passage under a similarity metric.

\section{Architecture}

The memory graph $G=(V,E)$ may contain goals, domains, episodes, events, facts, chunks, entities, procedures, decisions, sources, and agent actions. Typed edges may encode temporal order, causality, containment, contradiction, update, provenance, derivation, similarity, assignment, or dependency.

At runtime the system:
\begin{enumerate}
\item derives cues from the request, active goal, workflow, and agent state;
\item resolves authorized seed nodes and generates external-search candidates;
\item computes activation for candidate memory entities;
\item allocates a bounded traversal frontier using activation and fan attenuation;
\item expands graph paths, allowing retrieved entities to emit new cues;
\item stops under budget, marginal-evidence, contradiction, or coverage criteria;
\item reconstructs a typed evidence graph with provenance and unresolved gaps;
\item supplies the evidence graph to generation and records qualified outcomes.
\end{enumerate}

Authorization precedes content-bearing scoring and expansion. A node that is relevant but inaccessible must not leak content, identity, degree, or score through activation behavior. Concretely: the degree, decay, and co-occurrence statistics used in Equations~\ref{eq:base}--\ref{eq:fan-score} are computed over the requester's authorized subgraph only, so an inaccessible neighbor contributes to neither a node's reported degree nor its activation. Expansion tracks visited nodes per traversal and does not re-score or re-expand a node already in the frontier or evidence graph; a cycle in $G$ therefore shows up in redundancy accounting (Equation~\ref{eq:path-score}) rather than repeated work.

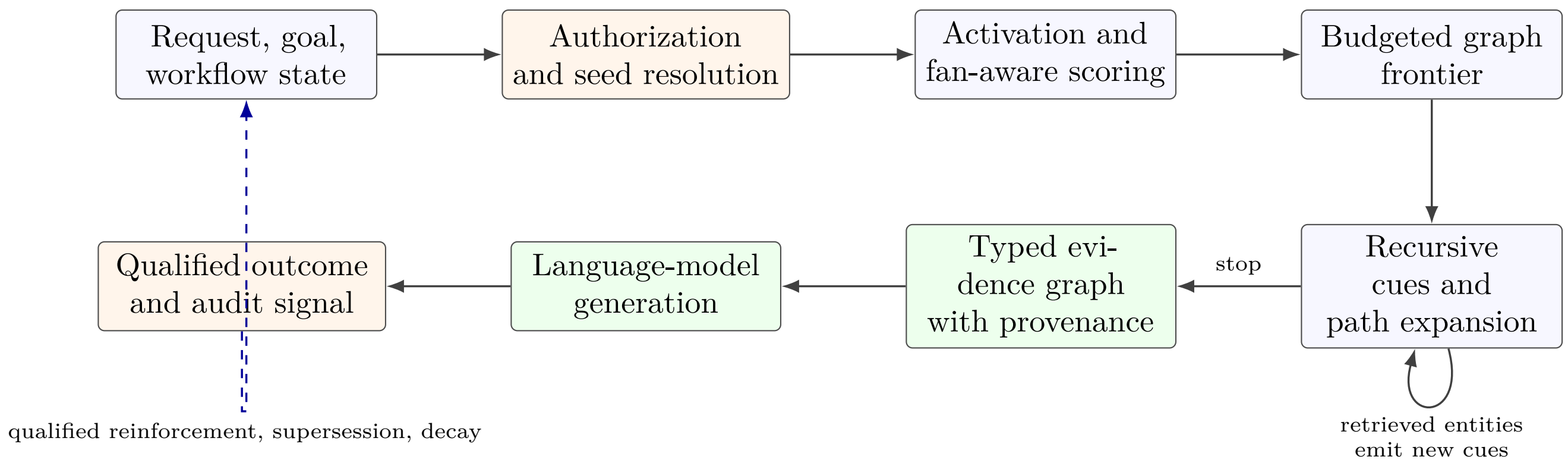
\begin{figure}[t]
\centering
\resizebox{\linewidth}{!}{%
\begin{tikzpicture}[
  node distance=14mm and 14mm,
  box/.style={draw=black!70, rounded corners=2pt, align=center, minimum height=10mm, text width=27mm, fill=blue!3},
  gate/.style={draw=black!70, rounded corners=2pt, align=center, minimum height=10mm, text width=30mm, fill=orange!8},
  output/.style={draw=black!70, rounded corners=2pt, align=center, minimum height=10mm, text width=28mm, fill=green!7},
  flow/.style={-{Latex[length=2mm]}, line width=0.65pt, draw=black!75},
  feedback/.style={-{Latex[length=2mm]}, line width=0.65pt, dashed, draw=blue!60!black}
]
\node[box] (input) {Request, goal,\\workflow state};
\node[gate, right=of input] (auth) {Authorization\\and seed resolution};
\node[box, right=of auth] (activation) {Activation and\\fan-aware scoring};
\node[box, right=of activation] (frontier) {Budgeted graph\\frontier};
\node[box, below=of frontier] (recursive) {Recursive cues and\\path expansion};
\node[output, left=of recursive] (evidence) {Typed evidence graph\\with provenance};
\node[output, left=of evidence] (generation) {Language-model\\generation};
\node[gate, left=of generation] (outcome) {Qualified outcome\\and audit signal};

\draw[flow] (input) -- (auth);
\draw[flow] (auth) -- (activation);
\draw[flow] (activation) -- (frontier);
\draw[flow] (frontier) -- (recursive);
\draw[flow] (recursive) to[loop below, looseness=6] node[below, font=\scriptsize, align=center] {retrieved entities\\emit new cues} (recursive);
\draw[flow] (recursive) -- node[above, font=\scriptsize] {stop} (evidence);
\draw[flow] (evidence) -- (generation);
\draw[flow] (generation) -- (outcome);
\draw[feedback] (outcome.south) -- ++(0,-9mm) -|
node[pos=0.35, below, font=\scriptsize] {qualified reinforcement, supersession, decay}
(input.south);
\end{tikzpicture}}
\caption{Activation-guided graph retrieval control loop. The self-loop at \emph{recursive cues and path expansion} is Section~\ref{sec:reconstruction}'s recursive cue generation: a retrieved entity's new cues re-enter frontier allocation before the stop criteria release the result to evidence reconstruction. Authorization is evaluated before content-bearing activation or expansion; only qualified downstream outcomes update future availability.}
\label{fig:architecture}
\end{figure}

\subsection{Activation model}\label{sec:activation}

For candidate $m_i$, a minimal base term is
\begin{equation}
B_i(t)=\log(n_i+1)-d_i\log(\Delta t_i+1)+\beta \kappa_i,
\label{eq:base}
\end{equation}
where $n_i$ is a qualified successful-use count, $d_i$ is a memory-specific decay rate, $\Delta t_i$ is elapsed time since a qualifying event, $\kappa_i$ is salience, and $\beta\ge 0$ weights salience. Equation \ref{eq:base} is an engineering approximation inspired by ACT-R base-level activation \cite{anderson1998atomic,anderson2004integrated}; it is not a cognitive-equivalence claim.

A runtime score may combine query relevance, spreading activation, utility, verification, and penalties:
\begin{equation}
A_i=\eta s_q(m_i)+B_i(t)+\sum_j W_jS_{ji}+U_i+Q_i-C_i+\epsilon.
\label{eq:activation}
\end{equation}
Here $\eta\ge 0$ weights query relevance, $S_{ji}$ is the association strength from cue $j$ to memory $i$ and $W_j$ the weight of cue $j$, $U_i$ is task utility, $Q_i$ is source or verification authority, $C_i$ penalizes contradiction, staleness, redundancy, or policy risk, and $\epsilon$ is a noise term, as in ACT-R, which a deterministic implementation may set to zero. Terms may be calibrated, learned, or replaced, but their contributions should remain auditable.

\subsection{Fan-aware traversal}\label{sec:fan}

High-degree cues such as ``customer,'' ``roadmap,'' or ``incident'' are weak discriminators. We regularize their traversal influence with
\begin{equation}
\phi(\deg(j))=\frac{1}{1+\alpha(\deg(j)-1)^2},
\label{eq:fan-attenuation}
\end{equation}
where $\alpha\ge 0$ sets the strength of attenuation, and score the move from cue $j$ to memory $i$ as
\begin{equation}
F_{ij}=W_j\,S_{ji}\,\phi(\deg(j))\,\psi(A_i),
\label{eq:fan-score}
\end{equation}
where $W_j$ and $S_{ji}$ are the cue weight and association strength from Equation~\ref{eq:activation}, and $\psi$ maps activation to traversal priority. A path score can be written
\begin{equation}
\operatorname{Score}(p)=\sum_{(j,i)\in p}F_{ij}
-\lambda\operatorname{Cost}(p)-\gamma\operatorname{Redundancy}(p),
\label{eq:path-score}
\end{equation}
where $p$ is a path of cue-to-memory moves and $\lambda,\gamma\ge 0$ weight its cost and redundancy.
The exact attenuation function is replaceable. The architectural requirement is that graph fan influences frontier allocation rather than being ignored.

\subsection{Recursive cues and evidence reconstruction}\label{sec:reconstruction}

Retrieved entities may introduce cues absent from the request. A cancellation record can activate a budget review; the review can activate a forecast; the forecast can activate a customer event. Recursive expansion continues while expected marginal evidence exceeds a threshold and the frontier remains within cost and authorization bounds.

The retrieval product is an evidence graph $G_e=(V_e,E_e)$, not an unordered list. $G_e$ retains selected evidence nodes, typed relations, source provenance, retrieval paths, activation contributions, confidence, contradictory branches, and unresolved gaps. Generation therefore receives explicit relational support rather than being required to infer all causal and temporal structure from adjacent text fragments.

\subsection{Worked Traversal}\label{sec:worked}

The following illustrates Sections~\ref{sec:activation}--\ref{sec:reconstruction} with representative values; no value here is measured.

\paragraph{Pseudocode.}
\begin{quote}
\small\ttfamily
frontier $\leftarrow$ seed(cues, $\pi$) \\
visited, evidence $\leftarrow$ $\emptyset$, $\emptyset$ \\
\textbf{while} frontier $\neq \emptyset$ and budget remains: \\
\hspace*{1.5em}$j \leftarrow$ pop highest-score cue from frontier \\
\hspace*{1.5em}\textbf{if} $j \in$ visited: \textbf{continue} \\
\hspace*{1.5em}visited $\leftarrow$ visited $\cup \{j\}$ \\
\hspace*{1.5em}candidates $\leftarrow$ authorized$(j, \pi)$ \\
\hspace*{1.5em}\textbf{for} $i$ in candidates: \\
\hspace*{3em}compute $A_i$ (Eq.~\ref{eq:activation}), $F_{ij}$ (Eq.~\ref{eq:fan-score}) \\
\hspace*{3em}\textbf{if} stop criteria met for $i$: evidence $\leftarrow$ evidence $\cup \{i\}$; \textbf{continue} \\
\hspace*{3em}frontier $\leftarrow$ frontier $\cup$ $(\text{cues}(i) \setminus \text{visited})$, scored by Eq.~\ref{eq:fan-score} \\
\textbf{return} evidence graph over evidence, visited, and recorded path scores
\end{quote}

\paragraph{Example.} Continuing the cancellation scenario of Section~\ref{sec:distinction}: the seed cue \emph{account loss} has $\deg=3$; with $\alpha=0.5$, $\phi(\deg)=1/(1+0.5\cdot2^2)=0.33$ (Equation~\ref{eq:fan-attenuation}). It reaches \emph{forecast revision} with association strength $S_{ji}=0.7$ and cue weight $W_j=1.0$, giving $F_{ij}=1.0\times0.7\times0.33\times\psi(A_i)\approx0.23\,\psi(A_i)$ (Equation~\ref{eq:fan-score}). \emph{Forecast revision} is retrieved and, under Section~\ref{sec:reconstruction}, emits the cue \emph{budget reduction}, which in turn emits \emph{cancellation}; the three hops' $F_{ij}$ values are recorded on the path and combined by Equation~\ref{eq:path-score}. A query-similarity reranker over the same initial candidate pool never introduces \emph{forecast revision} or \emph{budget reduction}, since neither shares vocabulary with the request; traversal reaches them because \emph{account loss} and \emph{forecast revision} emit them as cues after retrieval, not because they scored high against the original query. This is the specific contribution of recursive cue generation over propagating activation across a fixed, already-seeded neighborhood: the set of reachable nodes grows with what has already been retrieved, not only with graph distance from the original cues.

\section{Architectural Consequences}

The following consequences are argued from the specified architecture. They are not empirical performance claims.

\textbf{C1: Similarity is a candidate signal, not a complete memory policy.}
Query relevance omits goal state, use history, epistemic status, and path-dependent availability unless those variables are explicitly added.

\textbf{C2: Activation is most consequential as traversal control.}
Reranking a fixed top-$k$ pool cannot recover evidence excluded by initial candidate generation. Activation-guided expansion can allocate retrieval budget to graph paths whose downstream evidence is not query-similar.

\textbf{C3: Fan attenuation is graph regularization.}
Without degree-sensitive attenuation, generic high-degree cues can dominate expansion. Fan weighting reduces their frontier share while preserving them as routing signals.

\textbf{C4: Evidence reconstruction changes the retrieval contract.}
Returning typed relations, provenance, and unresolved gaps moves causal and temporal structure from implicit generation-time inference into an auditable retrieval artifact.

\textbf{C5: Reinforcement is a feedback-control problem.}
If retrieval increases future activation, false positives can become self-reinforcing. Updates must depend on qualified outcomes, and supersession must reduce authority without destroying lineage.

\section{Analytic Estimates}\label{sec:estimates}

All estimates in this section are implementation-level bounds under stated assumptions, not benchmark measurements.

\subsection{Traversal}

Naive breadth expansion with branching factor $b$ and depth $h$ visits $O(b^h)$ nodes in the worst case; $b$ here denotes a configured cap on neighbors expanded per node, not an empirical average, since an average bounds only the expected case, not the worst case. Suppose activation retains at most $L$ frontier nodes per depth and expands at most $b$ neighbors from each. Scoring examines at most $hLb$ candidate transitions. Maintaining a size-$L$ frontier with a heap yields
\begin{equation}
T_{\mathrm{traverse}}=O(hLb\log L),
\end{equation}
with $O(hLb)$ possible under linear-time bounded selection or bucketed scores. Memory for the active frontier is $O(Lb)$ if each depth is processed incrementally. Activation does not eliminate graph-search cost; it supplies a stateful pruning policy.

These bounds cover frontier selection and scoring only. If cue extraction (Section~\ref{sec:reconstruction}) calls a language model to derive cues from a newly retrieved entity, each such call adds its own latency and token cost on top of $T_{\mathrm{traverse}}$; a deterministic, non-model cue extractor keeps the stated bound exact, and a model-assisted extractor should be measured separately against it rather than assumed to fall within the same bound.

\subsection{Context and storage}

Flat retrieval of $k$ chunks averaging $c$ tokens costs approximately $kc$ context tokens. If the evidence graph retains $|V_e|$ node summaries averaging $u$ tokens, $|E_e|$ edge labels averaging $\ell$ tokens, and $\rho$ provenance tokens per node, then
\begin{equation}
T_{\mathrm{context}}\approx |V_e|(u+\rho)+|E_e|\ell.
\end{equation}
This quantity is not necessarily smaller than $kc$. The proposed benefit is higher evidence density and explicit relational support; evaluation must report both support coverage and token cost.

For $|V|$ memory entities and $|E|$ relations, graph storage is $O(|V|+|E|)$. Activation metadata adds $O(|V|)$ state plus optional access-history storage. A compressed or windowed event history is necessary if raw histories grow without bound.

\subsection{Reinforcement}

For the $|V_e|$ retrieved entities, post-retrieval metadata updates require $O(|V_e|)$ writes and can be asynchronous. The computational cost is small relative to model inference. The statistical risk is not: updating on retrieval alone implements a popularity loop. Qualified reinforcement should use later citation, confirmation, successful action, or consistency evidence.

\section{Evaluation and Falsification}\label{sec:evaluation}

We propose matched-model comparisons among (1) lexical retrieval, (2) dense or hybrid top-$k$, (3) graph retrieval without activation, (4) activation reranking over a fixed pool, (5) activation-guided traversal with recursive cues and evidence reconstruction, and (6) a write-time consolidated-notes baseline: a periodically rewritten, LLM-authored summary with a link index, built independently of query-time graph traversal. Anthropic reports such a system improving an internal evaluation's aggregate score from 53\% (search alone) to 61\% (search plus the consolidated notes) while reducing token cost, giving the falsification test in this section a named, public baseline to beat rather than only a hypothetical reranker \cite{anthropic2026memory}.

Tasks should require evidence structure: organizational why-questions, decision-history reconstruction, incident postmortems, temporal supersession, commitment recall, procedural next-action selection, and contradiction detection. Primary metrics are answer accuracy, evidence-node recall, causal-edge F1, temporal-order accuracy, contradiction recall, provenance completeness, stale-memory false-positive rate, and confidence calibration. Systems should also report latency, graph operations, context tokens, and human audit time.

Ablations remove base activation, goal state, fan attenuation, recursive cue generation, epistemic weighting, evidence reconstruction, reinforcement, and pre-retrieval authorization. The central claim is narrowed or falsified if a simpler graph reranker matches full traversal on causal and temporal recovery under equal model, evidence, and cost budgets.

\paragraph{Reproducibility protocol.}
Each comparison should publish the corpus snapshot or a synthetic generator, graph-construction rules, authorization policy, cue extractor, activation parameters, random seeds, model and prompt versions, traversal traces, retrieved evidence graphs, token and graph-operation counts, and per-task outputs. Statistical reporting should include paired confidence intervals, effect sizes, calibration error, and failure-case strata rather than aggregate accuracy alone. No benchmark dataset, implementation, or measured result is released with this architecture preprint; the protocol above defines the minimum artifact bundle for any subsequent empirical claim.

\section{Limitations, Failure Modes, and Scope}\label{sec:limits}

\paragraph{Limits of this paper.} The paper reports no measurements, so every benefit claimed for AGGR is a hypothesis until the evaluation in Section~\ref{sec:evaluation} is run. Equations~\ref{eq:base} and~\ref{eq:activation} are engineering approximations whose weights are uncalibrated; a deployment that sets them poorly can make activation worse than plain similarity. The architecture also assumes a memory graph whose nodes and typed edges are mostly correct: if extraction labels a budget review as \emph{caused by} a cancellation it merely preceded, traversal will reconstruct a confident but wrong causal chain.

\paragraph{Failure modes.} Activation creates state and therefore path dependence:
\begin{itemize}
\item \emph{Popularity lock-in.} Frequently retrieved memories can crowd out rare but decisive evidence, as when a much-cited status report keeps outranking the single customer email that explains a cancellation.
\item \emph{Stale authority.} Recency and repetition can preserve an obsolete decision. A pricing policy revised last quarter can keep its authority if the supersession edge carries too little weight.
\item \emph{Reflection laundering.} Model-generated summaries can acquire the authority of direct observations when lineage is lost, so an agent's guess about why a deal stalled is later retrieved as fact.
\item \emph{Fan overcorrection.} Strong attenuation can suppress broad cues that are genuinely diagnostic; for an outage question, ``incident'' may be the one cue that matters.
\item \emph{Goal capture.} An incorrect active goal steers traversal away from relevant evidence. A goal set to ``renewal'' can hide the support tickets that explain a churned account.
\item \emph{Permission leakage.} Scoring an inaccessible node before authorization can reveal that it exists, for example when a restricted HR record changes the ranking another user sees.
\item \emph{Audit overload.} An evidence graph can be inspectable in principle but too large to review in practice, such as a decision lineage of hundreds of nodes behind a one-line answer.
\end{itemize}
Mitigations require source lineage, explicit supersession, qualified reinforcement, authorization before scoring, calibrated stopping rules, and exposure of score contributions and path histories.

The architecture is unlikely to justify its overhead for short-lived factual question answering over a small, stable corpus. Its intended regime is long-lived agents and organizations where experience, goals, evidence authority, and access policy change over time.

\section{Related Work and Claim Boundary}

ACT-R establishes activation, spreading activation, declarative chunks, and procedural cognition \cite{anderson1998atomic,anderson2004integrated}. RAG establishes retrieval from non-parametric stores for generation \cite{lewis2020rag}. GraphRAG and RAPTOR establish graph- and hierarchy-structured retrieval \cite{edge2024graphrag,sarthi2024raptor}. Generative Agents, MemGPT, and temporal knowledge-graph memory establish persistent agent-memory mechanisms \cite{park2023generative,packer2023memgpt,rasmussen2025zep}.

Closer to this proposal, SYNAPSE models agent memory as a graph in which relevance emerges from spreading activation with lateral inhibition and degree-based attenuation, and reports gains on temporal and multi-hop recall \cite{jiang2026synapse}. Its activation step shares the fan-effect and degree-attenuation framing used here, but it runs a fixed number of propagation steps from lexical and dense-retrieval anchors and returns a ranked list of nodes; retrieved nodes do not introduce new cues, and no authorization step precedes scoring. HippoRAG similarly recovers multi-hop evidence from a knowledge graph, using Personalized PageRank from query-linked nodes rather than cue-driven traversal, and also returns a ranked list \cite{gutierrez2024hipporag}. A-MEM organizes agent memory through LLM-driven note construction, link generation, and periodic memory evolution \cite{xu2025amem}; memory organization, not retrieval-time control, is its central problem. SYNAPSE and HippoRAG, like this proposal, treat retrieval-time activation and multi-hop recovery as central, but neither lets a retrieved node introduce a new cue, neither authorizes before scoring, and both return a ranked list rather than a typed evidence graph.

We do not claim these components independently. What is new is the ordered runtime combination: activation that decides which memories become available rather than which records match, fan-aware traversal with recursive cues, evidence-graph reconstruction before inference, and reinforcement only after qualified outcomes. Each part has precedent; the combination is the contribution.

\section{Conclusion}

\enlargethispage{2\baselineskip}
Querying asks which stored objects match a request. Activation determines which prior experiences and evidence paths should be allowed to shape present behavior. The distinction turns retrieval from a stateless ranking step into a memory-control problem involving goals, history, graph structure, epistemic authority, authorization, and feedback. The architecture yields design consequences and tractable cost bounds; whether it improves model behavior is an empirical question with clear baselines, ablations, and falsification criteria.

\section*{Declarations}

\paragraph{Author contributions.}
Marvin Danig conceived the architecture, developed the analytic specification, and wrote the manuscript.

\paragraph{Funding.}
This research received no external funding.

\paragraph{Competing interests.}
The author is affiliated with Achiral Research. Certain aspects of the described architecture may be covered by patent applications associated with Achiral.

\paragraph{Data, code, and materials availability.}
This paper reports an architecture and analytic results and therefore introduces no experimental dataset or benchmark code. A companion essay, revision record, and future artifact links are maintained at
\href{https://achiral.ai/research/activation-guided-graph-retrieval}{\texttt{achiral.ai/research/activation-guided-graph-retrieval}}.

\paragraph{Ethics and limitations.}
The proposed system can amplify stale, biased, or incorrectly authorized memory if activation and reinforcement are poorly calibrated. Section~\ref{sec:limits} states the principal limitations and failure modes; deployment requires provenance, access-control testing, outcome-qualified updates, and auditable retrieval traces.

\end{document}